# Using Deep Learning Method for Classification: A Proposed Algorithm for the ISIC 2017 Skin Lesion Classification Challenge

Wenhao Zhang*, Liangcai Gao* and Runtao Liu*

*Institute of Computer Science & Technology, Peking University
Mar 5, 2017

## 1. Motivation

Skin cancer, the most common human malignancy, is primarily diagnosed visually by physicians [1]. Classification with an automated method like CNN [2, 3] shows potential for diagnosing the skin cancer according to the medical photographs[1]. By now, the deep convolutional neural networks can achieve the level of human dermatologist [1]. This work is dedicated on developing a Deep Learning method for ISIC [5] 2017 Skin Lesion Detection Competition[6] to classify the dermatology pictures, which is aiming at improving the diagnostic accuracy rate. As a result, it will improve the general level of the human health. The challenge falls into three sub-challenges, including Lesion Segmentation, Lesion Dermoscopic Feature Extraction and Lesion Classification. We focus on the Lesion Classification task.

The proposed algorithm is comprised of three steps: (1) original images preprocessing, (2) modelling the processed images using CNN [2, 3] in Caffe [4] framework, (3) predicting the test images and calculating the scores that represent the likelihood of corresponding classification. The models are built on the source images are using the Caffe [4] framework. The scores in prediction step are obtained by two different models from the source images.

## 2. The Architecture of the Classification Algorithm

The following sections describe the architecture of the proposed algorithm in details and show how to construct the models and final classification scores.

### 2.1 Image preprocessing

The image preprocessing is critical to subsequent steps and plays an important role in achieving a good performance. The source dermoscpic images provided by the competition are obtained from real clinic environment, which means that there are a great number of man-made interferences. The objective of preprocessing is to improve the quality of the source images, and make the following parts run easily.

### Image cropping

The interferences in the source images include medical auxiliary liquid, human hairs, clinical marks, etc. This section simply crops the irrelevant part of the source images, which is done manually and makes the images smaller.

### Size normalization

As a requirement of the CNN [2, 3] method and Caffe [4] framework, all the input images need to be normalized to 256*256. This section simply rescales the croped images to meet the requirement.

### Data augmentation

Requirement of huge amounts of data is one prominent characteristic of Deep Learning method. Without enough data, it's difficult to get a stable model and easily result in overfitting. As discussed above, expanding the scale of source images, which is also known as data augmentation, is of great importance. The algorithm deals with it through affine transforming the cropped and rescaled source images, like amplification, shrink, symmetry, etc.

## 2.2 Modelling the images

We learn a model with assistance of GPU in the Caffe [4] framework, which obtains the nonlinear and complicated relationship between the images and the labels The process of modelling the images is also called training the image data set.

## 2.3 Image classification

Through the well-trained model from the processed image set, Deep Learning method like CNN [2, 3] is capable to distinguish data of different classes. The model obtained from the processed images is trained by the combination of different parameters and fine-tuning.

## 2.4 Prediction of the classification

Algorithm generates two results for each test image. Responding to the requests of the competition, one should pertain the first binary classification task (melanoma vs. nevus and seborrheic keratosis) and the other should pertain the second binary classification task (seborrheic keratosis vs. melanoma and nevus). The algorithm prepares two copies of image data with different binary classification labels. In the training time, it runs on each training set and construct two different models. With these two models, the set of test images are able to be classified into two classes according to the probability, which is rescaled to [0.0, 1.0] with the following sigmoid conversion.

$$\frac{1}{e^{-a(x-b)}}$$

Where x is the original score, b is the binary threshold, and a is a scaling parameter.

## 3. The results and discussion

The result of algorithm, leaving some work haven't been done, is not very inspiring. Although an expansion of data has been done, the data scale is not big enough, which means there is no enough data to learn the internal relationships of source images and labels. Efforts in improving the algorithm will continue to go on. Following work will concentrate on these aspects: (1) employ more preprocess method to make the source images of higher quality, (2) try to use other machine learning methods to improve the classification accuracy.